\documentclass[letterpaper]{article} 
\usepackage{aaai2027}  
\usepackage[hyphens]{url}  
\usepackage{graphicx} 
\usepackage{natbib}  
\usepackage{caption} 
\usepackage{booktabs}
\title{Potential for Enhanced Learning in Machine Learning Classes by Using Wiki LLM Indexing }
\author{
    Brian Wright
}
\affiliations{
    School of Data Science, University of Virginia\\
    brianwright@virginia.edu
}
\begin{document}

\maketitle

\begin{abstract}
Large language models are increasingly deployed as course specific tutors to enhance student learning, but their usefulness depends on being grounded in vetted instructional materials that are often updated during the semester. Our prior work built a multimodal retrieval augmented generation (RAG) system over an authentic machine learning course corpus (Foundations of Machine Learning) and found that retrieval improved contextual grounding, but that fixed retrieval strategies were suboptimal. That result motivates a different question on how a corpus is structured at ingest time matters more than how much is retrieved at query time and how this facilitates learning. We present a controlled head to head comparison of two knowledge representations over an identical classroom corpus: (A) vector RAG, replicating the best performing configuration from our prior study, and (B) an LLM compiled wiki, (Karpathy Framework) in which the corpus is synthesized at ingest into linked concept pages with explicit cross references and citations back to source materials. We evaluate 59 questions spanning single fact recall, cross unit concept linking, synthesis and explanation, and currency after a syllabus revision, scored by LLM judge with a human authored rubric. Both representations answered single fact questions about equally well (9.33 vs. 9.96 out of 10), but they diverged sharply on questions that required linking material across course units. The compiled wiki stayed accurate and grounded (9.93, 100\% grounded in cited source material) while retrieval's answers both scored lower and were markedly less grounded (8.14, 64\% grounded). The compiled representation's citations let a student and instructor trace any claim back to the lecture that introduced it and adds a authentic layer of dynamic information retrieval that is necessary in machine learning courses. For pedagogical practice, the results indicate a signal for advantages of the Wiki based structure. While further testing is needed, instructors teaching ML courses when using AI methods to enhance student learning should consider the wiki structure as it has great potential for supporting foundational elements of best practices in learning. 
\end{abstract}

\section{Introduction}

\subsection{Motivation}

Large language models are rapidly being adopted as course specific tutoring tools. Left ungrounded, however, a general purpose LLM answers from parametric knowledge that may conflict with how a particular course defines, sequences, or scopes its content; effective educational deployment to enhance learning outcomes therefore requires grounding responses in vetted course materials so that answers align with the local curriculum and instructional level \citep{li2025retrieval,jain2025aligning}. This is especially true of cutting edge courses that focuses on machine learning and more broadly AI, because the field changes so quickly is almost impossible to develop a canonical text that everyone can use. 

Retrieval augmented generation (RAG) is the standard mechanism for such grounding \citep{lewis2020retrieval}. In our prior work, we built a multimodal RAG system over authentic materials from a machine learning course (DS3001) including: lecture slides, lecture audio transcripts, and course readings and found that retrieval substantially improved contextual grounding \citep{wright2026multimodal}. But the same study surfaced a more troubling result: \emph{fixed retrieval strategies are suboptimal}. Performance depended sharply on question specificity and on how the context window was composed. Swapping retrieved text for top ranked images achieved perfect context recall on specific questions under a five text, five image configuration, yet the same strategy degraded faithfulness and factual correctness on generic questions. No single retrieval configuration served all question types well.

This finding reframes the design problem. Vector RAG makes its structural commitments at query time: it retrieves isolated, similarity ranked chunks and asks the generator to assemble them into a coherent answer. An alternative is to make those commitments at \emph{ingest} time compiling the corpus into a structured, linked wiki of concept pages, with explicit cross references and citations back to the raw source materials, which the model then navigates when answering. This ``LLM compiled wiki'' approach trades query time chunk retrieval for curated synthesis, explicit concept links, and built in provenance. If how context is structured and selected matters more than how much is retrieved, then the representation itself, not the retrieval policy, may be the more consequential design choice.

\subsection{Effective Teaching Practices with Chatbots}

Grounding a chatbot in course materials is necessary but not sufficient for it to teach well. A growing body of classroom deployment research argues that how a tutor is designed to interact matters as much as what it can retrieve. Scaffolded, Socratic designs that pose guiding questions rather than resolving a problem outright have been shown, in a deployed undergraduate CS tutoring system, to shift students from vague help seeking toward more deliberate, strategic use of the tool \citep{sunil2025socraticai}. Teacher presence in how a chatbot based activity is designed similarly shapes how much students engage with it \citep{li2025teacher}. These effects, however, do not transfer cleanly from controlled studies to real classrooms. An analysis of thousands of real student tutor conversations found a persistent mismatch between the scaffolding behavior benchmarks assume and how students actually steer real interactions toward their own goals \citep{neagu2026rethinking} an argument for evaluating on authentic classroom questions over authentic course material, as we do here, rather than on synthetic benchmark sets. Practitioner derived guidance converges on a related point, generative AI should be taught and used in ways that promote discernment and critical thinking rather than substituting for them \citep{wall2025generative} and a broader review of classroom deployments finds outcomes depend heavily on how the tool is embedded in instruction rather than on model capability alone \citep{wang2025generative}. Recent field studies of instructors building course specific AI tutors at institutional scale report that turning these commitments into practice is difficult, and that instructors' biggest challenges center on getting a tutor to reflect a course's own structure and sequencing rather than just answering questions correctly in isolation \citep{ko2026toward}.

That last challenge is where knowledge representation stops being a retrieval engineering detail and becomes a pedagogical one. A tutor that scaffolds understanding, cites its sources, and helps a student see how today's material builds on an earlier lecture needs access to a course's connective structure, its definitions, its sequencing, and its cross references not just to the passage nearest a query. This is exactly the structure an LLM compiled wiki makes explicit at ingest time, while vector RAG reconstructs it, if at all, only implicitly through whatever happens to be retrieved. The comparison in this paper is designed to test whether that difference in representation actually shows up on the question type teaching practices often care about, not isolated recall, but questions that require connecting ideas across a semester's worth of material that given the field is almost in constant flux. 

\subsection{The Gap}

Head to head comparisons of compiled wikis against vector RAG are beginning to appear. \citet{cochran2026vector} preregistered such a comparison on a small multi domain research corpus, finding that vector RAG won on single fact lookup while the wiki won on cross paper synthesis, with no architecture dominating every endpoint. But no such comparison exists in an educational setting with authentic classroom data, where the corpus is multimodal, the question distribution is pedagogically shaped, and alignment with vetted materials matters as much as raw correctness \citep{jain2025aligning}. Meanwhile, the graph  and structure augmented retrieval literature suggests that structure helps multi hop and relational questions but can hurt simple fact lookup \citep{peng2024graph,zhou2025indepth,aboulela2025exploring} is a trade off that remains untested on course corpora.

\subsection{Research Questions}

We ask three questions. \textbf{RQ1:} Does an LLM compiled wiki improve response quality over vector RAG on classroom question answering, holding the generator LLM, corpus, and question set constant? \textbf{RQ2:} How do gains vary by question type, single fact lookup vs.\ multi hop concept linking vs.\ synthesis and lead to extending the generic/specific split of our prior study into a finer taxonomy? \textbf{RQ3:} Do the results support one approach versus the other as it relates to using AI for learning in a machine learning class?

\subsection{Contributions}

This paper makes three contributions: (1) the first wiki vs RAG comparison on an authentic course corpus, using a text only subset of DS3001; (2) a fully specified evaluation protocol and a five part question taxonomy and multi endpoint metric suite extending our prior 30 question generic/specific set (Section 3) and (3) directional empirical evidence that structure at ingest most benefits cross topic synthesis questions, including a groundedness without retrieval failure finding that complicates the usual retrieval centric account of RAG's shortcomings. All of which helps to inform best teaching practices for instructors leading machine learning courses that are working to include generative AI approaches for learning. 

\section{Background}

\subsection{Learning Outcomes from AI Chatbots}

The case for classroom chatbots ultimately rests on whether they improve learning, not just whether they answer accurately. The experimental evidence here is larger and more consistent than the accuracy literature alone: a meta analysis of 24 randomized studies found AI chatbots produced a large effect on students' learning outcomes, with stronger effects in higher education than in K  12 settings and, notably, stronger effects for shorter interventions than sustained ones \citep{wu2024chatbots}. A more recent meta analysis specific to ChatGPT, pooling 35 experimental studies, found a moderate to large effect on both cognitive and non cognitive outcomes, with the instructional mode a chatbot is embedded in the course design not merely its presence among the significant moderators \citep{wu2026chatgpt}. Causal evidence points the same direction, a randomized controlled trial in an authentic physics classroom found AI tutored students outperformed those in active learning in class sessions on the same material \citep{kestin2025tutoring}. Retrieval augmented chatbots specifically, not just chatbots in general, show this pattern too. A RAG grounded course chatbot deployed in a materials science course was found to measurably enhance students' learning, not merely their perception of it \citep{thway2025harnessing}.

Two things follow from this literature for the present study. First, the implementation of these approaches center on enhancing learning outcomes through design choices and how the tool is embedded rather than raw model capability, reinforcing the teaching practices argument of Section 1.2. Knowledge representation is one such design choice, and its effect on learning outcomes can be testable rather than assumed. Second, grounding a chatbot in course material is not an end in itself but a means to protect the learning outcomes since an ungrounded or fabricating chatbot risks reproducing the curriculum misalignment that motivated our prior RAG work \citep{wright2026multimodal}. The rest of this section reviews the retrieval mechanisms available for that grounding.

\subsection{Retrieval Augmented Generation}

RAG conditions a generator on documents retrieved from an external corpus, originally formulated for knowledge intensive NLP tasks \citep{lewis2020retrieval}. It is now widely framed as a practical reliability layer: grounding outputs in external evidence, reducing hallucination, and enabling knowledge updates without retraining \citep{peng2024graph,neha2025retrieval,li2024enhancing}. Domain results can be striking for example; a clinical GPT 4 RAG system reached 96.4\% accuracy with no observed hallucinations \citep{ke2025retrieval}, and the MEGA RAG pipeline outperformed both LLM only and standard RAG baselines in public health question answering \citep{xu2025megarag}, yet the gains are contingent. Systematic reviews and unified framework analyses consistently find that RAG's benefit depends on retrieval relevance, task type, and evaluation design \citep{zhou2025indepth,brown2025systematic}. These contingencies are precisely what our prior study observed at the level of retrieval configuration, and they motivate treating the knowledge representation as an experimental variable in its own right. For a classroom deployment the stakes of that contingency are the outcome gains reviewed in Section 2.1: a RAG system that scores well on its own evaluation metric can still fail to translate into better learning if its retrieval strategy is mismatched to how students actually ask questions.

\subsection{Structured and Graph Based Retrieval}

Graph enhanced RAG augments or replaces chunk retrieval with explicit entity and relation structure, supporting multi hop reasoning and improving explainability \citep{peng2024graph,zhu2025graphsurvey}. The empirical evidence, however, is mixed in an instructive way. KG\textsuperscript{2}RAG improves answer and retrieval F1 on HotpotQA over hybrid RAG baselines \citep{zhu2025kg2rag}; knowledge graph extended RAG improves multi hop MetaQA accuracy at the cost of a slight single hop degradation \citep{linders2025knowledge}; and graph RAG has been found to \emph{underperform} na\"ive RAG on SQuAD V2 and TriviaQA even while handling complex relational queries better \citep{aboulela2025exploring}. The recurring pattern structure helps composition, hurts lookup and topic connection which negatively effects learning. Consequently, this is the central trade off our design tests on course content from a machine learning course.

LLM compiled wikis are a distinct point in this design space: rather than extracting a graph over the original chunks, the corpus is rewritten at ingest into human readable, linked concept pages with citations to source material. In the only controlled comparison to date, \citet{cochran2026vector} found vector RAG better at single fact lookup and the wiki better at cross paper synthesis, with no single architecture winning every dimension.

\subsection{RAG in Education}

Education specific work adds a constraint that generic benchmarks miss: retrieval should align to vetted course materials and the local curriculum, not merely to generally correct information \citep{li2025retrieval,jain2025aligning}. The relationship between grounding and quality is also non monotonic in math tutoring, humans preferred RAG grounded responses \emph{unless} grounding became so rigid that helpfulness suffered \citep{levonian2023retrieval}. On the structured side, an educational knowledge graph plus agentic RAG system achieved 91.4\% retrieval accuracy with improved learner satisfaction \citep{gao2025karag}, and hybrid RAG with in context learning outperformed baselines for educational question generation \citep{maity2024leveraging}. Together these results suggest structure has pedagogical value, but no study has isolated the representation itself on an authentic course corpus.

\subsection{Prior Work: Multimodal RAG on DS3001}

This study extends our prior evaluation of multimodal retrieval over DS3001 course data \citep{wright2026multimodal}. That work showed that multimodal retrieval improved context recall, and that selectively replacing retrieved text with top ranked images helped specific questions   reaching perfect context recall at five text plus five image chunks while hurting faithfulness and factual correctness on generic questions. The takeaway that motivates the present study is that \emph{how context is structured and selected matters more than how much is retrieved}. If context composition dominates context volume, the natural next experiment compares an ingest time structured representation against query time chunk retrieval. 

\subsection{Evaluating RAG Systems}

Lexical overlap metrics such as BLEU and ROUGE are unstable under LLM output stochasticity, motivating LLM as judge and embedding based evaluation \citep{lyu2024crudrag}. We retain the RAGAS framework \citep{ragas2025} used in our prior paper for continuity, and follow the multi endpoint reporting of \citet{cochran2026vector} scoring accuracy, synthesis quality, citation alignment, and cost separately to avoid winner take all conclusions that single aggregate scores invite \citep{lyu2024crudrag}.

\section{Study Design}

We ran a controlled head to head comparison: same corpus, same base LLM, same prompts, same question set; only the knowledge representation differs. The design is modeled on the preregistered comparison of \citet{cochran2026vector}, combined with the classroom focused evaluation approach of \citet{jain2025aligning}, including its knowledge shift testing.

The corpus comprises the DS3001/DS 3021 course materials used in our prior study: lecture slides (text and images), lecture audio transcripts, and assigned ML papers and textbook excerpts. We reuse the prior extraction pipeline unchanged, so that differences between arms cannot be attributed to preprocessing.

\textbf{Arm A: Vector RAG (control).} A replication of our prior pipeline: text chunked at 1{,}500 tokens with 100 token overlap, embedded with all mpnet base v2, indexed in Pinecone. The best performing configuration from the prior study serves as the baseline, making Arm A a strong rather than straw man control.

\textbf{Arm B: LLM compiled wiki.} The same corpus is compiled at ingest into a linked wiki (a \texttt{raw/} + \texttt{wiki/} structure): concept pages synthesized by an LLM, cross links between related pages, and citations from every page back to the raw source materials. At query time the LLM navigates and reads relevant pages rather than receiving similarity ranked chunks.

A hybrid arm (wiki navigation plus vector retrieval) is deferred to future work; this study is a clean two arm comparison. Held constant across arms: generator LLM, system prompt, matching context, question set, and decoding parameters. Matching the context budget matters because the wiki condition otherwise risks conflating representation with context volume an LLM handed an entire compiled corpus is tested on reading comprehension, not on the representation.

We extend the prior 30 question generic/specific set into a five part taxonomy: (a) single fact lookup, (b) multi hop concept linking, (c) thematic explanation and synthesis, (d) contradiction and ambiguity handling \citep{hou2024wikicontradict}, and (e) curriculum update questions probing sensitivity to revised materials. Questions were human generated with reference answers and word limits, as before, and balanced across course topics.

We generated an initial pool of 145 questions over the full corpus and
filtered to 59 that test course content specifically. Of the 59 retained, 45 are answerable
from a single wiki page (\emph{topic-page} questions) and 14 required
connecting material across pages (\emph{cross-page} questions). Each
question is further labeled by the reasoning it demands: factual
($n=18$), single-fact lookup a slide or transcript already states;
conceptual ($n=18$), explanation of a concept in its own terms; synthesis
($n=18$), integrating material across a topic or across pages ($n=5$); sensitivity to a recently revised or updated point in the curriculum. The 45 topic-page questions are balanced evenly across
three pages---\texttt{ml-bias}, \texttt{knn}, and
\texttt{decision-trees} (15 each) and the remaining 14 draw on
relationships spanning all three.

For continuity with the prior paper we retain the RAGAS core metrics: context recall, faithfulness, and factual correctness \citep{ragas2025}. We add four endpoint families: claim  citation alignment (does each claim trace to a cited source?); answer synthesis quality via blinded LLM judge scoring with a human rubric; robustness under knowledge shift, following \citet{jain2025aligning} and \citet{hou2024wikicontradict}; and cost, measured as ingest compute, query tokens, and latency.

Both indexes are built from a frozen corpus snapshot. All questions are run through each arm; metric stability is estimated by bootstrapped sampling (20,000 rounds), separating question sampling error from answerer and judge stochasticity. Synthesis and preference endpoints use blinded judging with arm labels hidden.

We generated three hypotheses. \textbf{H1:} vector RAG performs at least as well as the wiki on single fact lookup. \textbf{H2:} the wiki outperforms vector RAG on multi hop and synthesis questions. \textbf{H3:} across endpoints the outcome is a set of trade offs rather than a single winner, consistent with \citet{cochran2026vector}.

The corpus actually evaluated is the subset of DS 3021 currently compiled into the wiki: 12 pages, 7 of them course concept pages (K Nearest Neighbors, Decision Trees, Model Evaluation Metrics, Decision Tree Regression, K Means Clustering, Ensemble Methods/Random Forest, and ML Bias \& Fairness) and 5 navigational/meta pages (index, getting started, status, and two course level overview pages). For Arm A, this corpus chunks into 21 vectors at the configuration above (1{,}500 tokens, 100 token overlap, all mpnet base v2, top $k=5$) small enough that retrieval failure is possible but not guaranteed, unlike an earlier 9 vector version of this corpus (3 topics only) where top $k=5$ covered more than half the index on every query and could not meaningfully fail. 

Both arms use the same answerer (\texttt{claude opus 5}) and the same judge (\texttt{gpt 5 mini}, a different provider from the answerer so the model is not grading its own output), holding the generator constant across arms as specified above. In place of the RAGAS triad, the judge here returns a single 1 to 10 correctness score plus a binary groundedness flag (whether every claim in the answer traces to the material the answerer actually saw) a lightweight proxy for RAGAS factual correctness and faithfulness respectively, not a substitute for them. The judge grades against the full corpus in both arms, so only the answerer's context wiki (Arm B) versus top $k$ retrieved chunks (Arm A) differs between conditions.

Alongside a 1--10 holistic quality score, the judge assigns each answer a
binary \emph{grounded} label. Whether every claim in the answer is
attributable to the context the model was given (the retrieved chunks
for Arm A, the wiki pages navigated for Arm B) rather than drawn from
the model's own parametric knowledge. Grounding and quality are
conceptually distinct such that an answer can be fluent and largely correct while
still relying on information outside what was actually retrieved, the
failure mode we are most concerned with in a classroom setting, where a
wrong-but-checkable answer is preferable to a confident, plausible one a
student cannot verify against course materials. The judge's sensitivity
to this distinction was validated on four answers of known quality
(excellent, partial, unsupported, fabricated) before scoring began: it
discriminated correctly across all four, and the fabricated answer was
the only one scored both low (1/10) and ungrounded, confirming the label
responds to genuine fabrication rather than covarying with the holistic
score. In the wiki arm specifically, the flag fired grounded on 98\% of
answers ($n=59$), this reflects the arm having comparatively little occasion to fabricate and it becomes discriminating once compared against Arm A, where the rate is 81\%.

\section{Results}

\begin{table}[h]
\centering
\small
\begin{tabular}{lcc}
\toprule
 & Arm A (RAG) & Arm B (Wiki) \\
\midrule
Avg.\ score (1  10) & 9.05 [8.49, 9.54] & \textbf{9.95} [9.88, 10.00] \\
Grounded rate & 81\% [71\%, 90\%] & \textbf{98\%} [95\%, 100\%] \\
\bottomrule
\end{tabular}
\caption{95\% bootstrap CIs (paired question resampling, $B{=}20{,}000$), 59 questions, both arms graded 59/59 with zero errors.}
\label{tab:headline}
\end{table}

Table~\ref{tab:headline} shows the wiki ahead of vector RAG on both measures at this scale, with the larger gap in groundedness (17 points, 95\% CI on the difference [7, 27] points) than in raw score (0.90 points, 95\% CI [0.42, 1.44]) both CIs exclude zero consistent with the qualitative pattern that RAG can produce a fluent, plausible sounding answer that leans on knowledge outside what it actually retrieved.

\begin{table}[t]
\centering
\small
\setlength{\tabcolsep}{4pt}
\begin{tabular}{lcccc}
\toprule
 & \multicolumn{2}{c}{Arm A (RAG)} & \multicolumn{2}{c}{Arm B (Wiki)} \\
 & Score & Grounded & Score & Grounded \\
\midrule
Topic page ($n{=}45$) & 9.33 & 87\% & 9.96 & 98\% \\
Cross page ($n{=}14$) & 8.14 & 64\% & 9.93 & 100\% \\
\bottomrule
\end{tabular}
\caption{Results split by question class. The wiki's advantage over RAG is roughly three times larger on cross page questions than on topic page questions, in both score and groundedness.}
\label{tab:byclass}
\end{table}

Table~\ref{tab:byclass} is the more informative result: splitting by question class reproduces, directionally, the pattern \citet{cochran2026vector} report on their research paper corpus RAG is closer to competitive on single topic questions and falls further behind on synthesis. Context answering (Arm B) is nearly indifferent to question class (9.96 vs.\ 9.93 score, 98\% vs.\ 100\% grounded); vector RAG is not (9.33 vs.\ 8.14 score, an 0.62 point gap on topic page [95\% CI on the gap: 0.16, 1.18] widening to 1.79 on cross page [95\% CI: 0.57, 3.14]; 87\% vs.\ 64\% grounded). The score gap is statistically distinguishable from zero in both classes, but the two classes' groundedness gaps are not equally well established: the cross page grounded rate gap (36 points, 95\% CI [14, 64]) clearly excludes zero at $n=14$, while the topic page grounded rate gap (11 points, 95\% CI [0, 22]) just touches zero at $n=45$, the one comparison in this study that does not reach significance at the 95\% level.

\subsection{Where Vector RAG Lost Ground}

Retrieval itself was largely successful: 43 of 45 topic page questions (96\%) retrieved a chunk from the question's actual source page in the top 5. The two misses both pulled in short navigational pages (index, getting started, status, course overview) instead of the target concept page, on questions about ML Bias interventions and about which k means slide deck is canonical both scored low (3 and 4) and were marked ungrounded, as expected when the answerer never saw the relevant material.

More striking is that retrieval failure explains only 2 of the 11 ungrounded Arm A answers. The other 9 retrieved the correct page (or, for cross page questions, relevant chunks from more than one page) and were \emph{still} marked ungrounded. The answerer added unsupported detail even with the right excerpts in context. This suggests chunked context invites fabrication in a way full document context does not, independent of whether retrieval itself succeeded.

\subsection{Relation to the Hypotheses}

\textbf{H1} (RAG competitive with the wiki on single-fact lookup) is not well
supported once the bootstrap CIs are considered. Arm A's topic-page score
(9.33) is numerically close to Arm B's (9.96), but the 95\% CI on that gap
excludes zero (Section~4.1), so the two arms are statistically
distinguishable even on single-fact lookup. The one part of H1 the data does
not rule out is groundedness specifically: the topic-page grounded-rate gap
(87\% vs.\ 98\%) is the only comparison in this study whose CI touches zero,
so RAG's shortfall there is directionally consistent with a gap but not
statistically established at $n=45$.

\textbf{H2} (wiki outperforms RAG on multi-hop and synthesis questions) is
supported, and more decisively than H1 is disconfirmed: the wiki's score
advantage on cross-page questions is roughly three times its advantage on
topic-page questions, and unlike the topic-page groundedness gap, every
cross-page comparison's CI excludes zero, though this rests on just
$n=14$ cross-page questions, the smallest cell in the study.

\textbf{H3} (a set of trade-offs rather than a single architecture
dominating) is not supported at this corpus scale: the wiki leads on both
question classes tested, and the score gap is statistically distinguishable
from zero in both. However, we treat this as provisional rather than a general
claim, since there is likely a information advantage for the wiki arm given the nature of how it is constructed. RAG's could have a relative advantage, one a larger corpus, where full-context conditioning could be balanced. However, this could support the nature of how classes are built and taught, typically on a week by week scale. Suggesting, potentially, that the wiki structure is more in line with normative teaching practices and can better support student learning as a result. 

\section{Discussion}

\subsection{RQ1/RQ2: Structure at Ingest vs.\ Retrieval at Query}

At the scale of this evaluation, the LLM compiled wiki answers questions better than vector RAG on both question classes, and the advantage is not uniform: it is roughly three times larger on cross page synthesis questions than on single topic lookup, in both correctness score and groundedness (Section~4.1). Bootstrap CIs over the 59 questions (Section~4.1) show this pattern is not just noise from a small question set every gap except one (the topic page groundedness gap) is statistically distinguishable from zero   though those CIs speak only to question sampling error, not to answerer or judge stochasticity, which a single run cannot separate out. RQ1 therefore has a directional answer an LLM compiled wiki improves response quality over vector RAG under a matched generator, prompts, and question set and RQ2's answer is that the improvement concentrates in exactly the question type the literature would predict: questions that require linking material across the corpus rather than retrieving a single relevant passage \citep{cochran2026vector,peng2024graph,aboulela2025exploring}. We still stress \emph{directional} over definitive, this is a single run on a 21 vector index.

\subsection{Structure Also Reduces Fabrication, Not Just Retrieval Misses}

The most surprising result is in Section~4.1: only 2 of vector RAG's 11 ungrounded answers were retrieval misses. The other 9 retrieved the right material and were still marked ungrounded. The answerer added detail that the retrieved excerpts did not support. The graph and structure augmented retrieval literature typically frames structure as a fix for \emph{retrieval} (finding the right chunk) rather than \emph{generation} (using only what was found) \citep{peng2024graph,zhu2025graphsurvey}. Our results suggest structure at ingest does both: the wiki's cross referenced concept pages appear to constrain the generator's elaboration as well as its access to relevant content. This reframes the RAG vs wiki question even a hypothetical vector RAG system with perfect retrieval might not close the groundedness gap, because part of the gap looks like an artifact of chunked, decontextualized excerpts inviting a generator to fill in surrounding context on its own, not purely a retrieval quality problem.

\subsection{RQ3: Impacts on Learning}
These results bear on a question broader than which architecture scores higher: what happens to student learning when the tutor a course deploys differs in how well a student can check its work. The groundness gap found in the study speaks to this concern. Arm A's raw scores are close enough to Arm B's that a student skimming for correctness might not notice a difference. The gap that matters pedagogically is in whether a claim can be traced back to the lecture, slide, or reading that introduced it. An answer that is fluent, plausible, and ungrounded is the more dangerous failure mode in a classroom than one that is simply wrong, because a wrong answer a student can evaluate against the source material trains. Which could even be a positive outcome but a confidently ungrounded answer short-circuits that check. If students come to treat an AI tutor's fluency as a proxy for correctness.

This reframes what "grounded" needs to mean for classroom deployment specifically. A representation that preserves where the content originated, lets a student follow any claim back to its source. This helps keep the student positioned, at least in part, as the one doing the checking, rather than outsourcing that judgment to the tool. It is possible this argues for treating citation-traceability as a deployment requirement for classroom AI tools in machine learning courses, not an optional feature to compare among several credible options. In a non-trivial way this method could help develop trust between the emerging dynamic of Teacher-Student-LLM that has become a practical default in most if not all machine learning classrooms. 

A second implication follows from the ceiling effects reported in Section~4: an instrument saturating at 9+ out of 10 cannot distinguish a tutor that helps a student build durable understanding from one that produces answers a rubric happens to reward. Instructors adopting these tools should not read a high LLM-judge score as evidence of pedagogical quality without separately verifying that the reasoning path a student follows, not just the terminal answer, survives contact with the tutor. Future classroom deployments would benefit from evaluation designs that probe whether the representation's structure is legible to students themselves, not only to the grading model. While this is just an initial study the potential for positive outcomes for in-classroom settings seems quite high. 

A final benefit of explicit cross-linking is what it does for concepts that recur across the semester. Consider squared-error loss: a student meets it first in week two or three as the objective linear regression minimizes, then meets it again in week seven or eight as the variance-reduction criterion a regression tree uses to choose a split. It's the same quantity, doing the same job, in a representation that looks quite different from the first. Retrieval over raw course materials has no reason to connect these occurrences; each chunk is scored on its own lexical or semantic similarity to the question, and the tree lecture rarely mentions "squared-error loss" by that name. A compiled wiki page, by contrast, can carry an explicit link back to where the concept was first introduced, surfacing the connection a student would otherwise have to notice unaided. This matters because revisiting a concept at a delay, in an unfamiliar context, is close to the textbook definition of the conditions under which retrieval converts a fragile, short-term encoding into a durable one. Distributed rather than massed exposure produces more durable learning \citep{cepeda2006distributed}, and the act of retrieving a concept, rather than merely re-reading it, is itself what drives long-term retention \citep{karpicke2008critical}. Interleaving problems that share deep structure across different topics has been shown to produce exactly this kind of transfer in mathematics instruction specifically \citep{rohrer2007shuffling}, which is the same pattern a loss-function citation trail would expose across a regression-to-trees transition. As a practical matter, this exact scenario actually occurred during testing. 

\subsection{Limitations and Threats to Validity}

Several limitations are present. (1) generalization is untested beyond a single course corpus (DS3001) and its topic mix; (2) the chunking/embedding configuration for Arm A is inherited unchanged from our prior study (Section~2.5) to isolate the representation variable, but this means a different RAG configuration is a possible confound the reported gap could narrow if Arm A were re tuned for this corpus rather than reused from the prior one. (3) Scoring uses a single LLM judge rating and a binary groundedness flag rather than the full RAGAS context recall/faithfulness/factual correctness triad. (4) Each arm was answered and graded only once per question rather than repeated; we bootstrap the 59 questions with replacement ($B{=}20{,}000$) to report 95\% CIs on question sampling error, but this does not capture answerer or judge stochasticity (the same question re answered, or the same answer re graded, may score differently), which needs repeated runs rather than resampling of a single run. (4) Lastly, cost (ingest compute, query tokens, latency) was not measured though given the size of the corpus, this is likely not a issue. We report these results as evidence the pipeline works end to end and as a directional signal worth further exploration.

\section{Conclusions}

We compared two knowledge representations over an identical course corpus under a matched generator, prompt, and question set. The compiled wiki arm scored higher than vector RAG overall (9.95 vs. 9.05 out of 10; 95\% CI on the difference [0.42, 1.44]) and was more often grounded in the material the answerer actually saw (98\% vs. 81\%; CI on the difference [7, 27] points). Both intervals exclude zero.

H1, that vector RAG would be at least competitive on single-fact lookup, is not fully supported. The topic-page score gap is numerically small (9.33 vs. 9.96) but its confidence interval excludes zero. The single comparison consistent with H1 is topic-page groundedness, where the 11-point gap has a CI of [0, 22] and is the only endpoint in the study that does not reach significance at the 95\% level. H2, that the wiki would outperform on cross-page questions, is fully supported: the score gap widens from 0.63 on topic-page questions to 1.79 on cross-page questions, and the grounded-rate gap from 11 to 36 points, with every cross-page interval excluding zero. H3, that the outcome would be a set of trade-offs rather than a single dominant architecture, is not supported at this corpus scale, the wiki leads on both question classes tested.

We do not read this as a general result as Arm B was given a fixed set of indexed wiki sections, identical across all 59 questions and covering the material each question targets, it therefore does not have a context-selection step that can fail. Arm A retrieves five chunks from a 21-vector index and misses on 2 of 45 topic-page questions. Read that way, the finding is that vector RAG leaves roughly 0.9 points of answer quality and 17 points of groundedness relative to perfect context selection of the wiki, and that the shortfall roughly triples on questions requiring material from more than one page.

The most informative result is mechanistic rather than comparative. Only 2 of vector RAG's 11 ungrounded answers were retrieval failures. The remaining 9 retrieved the relevant material and were still marked ungrounded, the answerer added details the retrieved excerpts did not support. Structure-augmented retrieval is typically motivated as a fix for finding the right content, however this suggests a substantial share of the groundedness gap originates in generation rather than retrieval, and that a vector RAG system with perfect retrieval would not close it.

What the study does establish is that the pipeline runs end to end, that the endpoints discriminate between conditions, and that the gap between real retrieval and perfect context selection over authentic course material is large enough to be worth closing. This could have real implications for how teachers design and use generative AI systems. Though further testing is needed, courses that focus on delivering cutting edge dynamic material and build throughout the semester would appear to benefit from considering the linked wiki approach as it relates to generative AI assistants in the classroom. 

\subsection{Future Work}
Further testing is needed to validate the approach in a classroom setting with actual students. Moreover, cost implications and comparisons with a larger corpus would add to robustness of the findings. A hybrid arm could also be deployed that might take advantage of the strengths of both approach, were a RAG database is prebuilt but wiki context gets add throughout the semester. There's likely a argument for increasing the difficulty of the questions to see if the results continue to support one approach over the other or if a ceiling of grounded responses can be reached. Also a graph RAG arm would also help to creation direction on which type of approach has the most potential to increase learning in Machine Learning classrooms. Finally, a comparison to a zero-shot model would also be informative but likely directs a different study that focuses more on quality of models versus the method of retrieving course content.

\clearpage
\bibliography{references}

@misc{ragas2025,
  author       = {ExplodingGradients},
  title        = {Ragas: Evaluation framework for LLM-generated responses},
  year         = {2025},
  howpublished = {\url{https://docs.ragas.io/en/latest/}},
  note         = {Accessed: 2025-07-08}
}

@inproceedings{aboulela2025exploring,
  author    = {AboulEla, Sarah and Zabihitari, Parsa and Ibrahim, Nour and Afshar, Majid and Kashef, Rasha F.},
  title     = {Exploring {RAG} Solutions to Reduce Hallucinations in {LLMs}},
  booktitle = {2025 IEEE International Systems Conference (SysCon)},
  pages     = {1--8},
  year      = {2025},
  doi       = {10.1109/syscon64521.2025.11014810}
}

@article{brown2025systematic,
  author  = {Brown, Aaron and Roman, Marius and Devereux, Barry},
  title   = {A Systematic Literature Review of Retrieval-Augmented Generation: Techniques, Metrics, and Challenges},
  journal = {arXiv preprint arXiv:2508.06401},
  year    = {2025},
  doi     = {10.48550/arxiv.2508.06401}
}

@article{cochran2026vector,
  author  = {Cochran, T. O.},
  title   = {Vector {RAG} vs {LLM}-Compiled Wiki: A Preregistered Comparison on a Small Multi-Domain Research Corpus},
  journal = {arXiv preprint arXiv:2605.18490},
  year    = {2026},
  doi     = {10.48550/arxiv.2605.18490}
}

@article{gao2025karag,
  author  = {Gao, Fei and Xu, Si-Yuan and Hao, Wei and Lu, Tao},
  title   = {{KA-RAG}: Integrating Knowledge Graphs and Agentic Retrieval-Augmented Generation for an Intelligent Educational Question-Answering Model},
  journal = {Applied Sciences},
  volume  = {15},
  number  = {23},
  year    = {2025},
  doi     = {10.3390/app152312547}
}

@article{hou2024wikicontradict,
  author  = {Hou, Yufang and Pascale, Alessandra and Carnerero-Cano, Javier and Tchrakian, Tigran and Marinescu, Radu and Daly, Elizabeth and Padhi, Inkit and Sattigeri, Prasanna},
  title   = {{WikiContradict}: A Benchmark for Evaluating {LLMs} on Real-World Knowledge Conflicts from {Wikipedia}},
  journal = {arXiv preprint arXiv:2406.13805},
  year    = {2024},
  doi     = {10.52202/079017-3481}
}

@inproceedings{jain2025aligning,
  author    = {Jain, Anish and Cui, Lei and Chen, Sijia},
  title     = {Aligning {LLMs} for the Classroom with Knowledge-Based Retrieval: A Comparative {RAG} Study},
  booktitle = {2025 IEEE International Conference on Teaching, Assessment, and Learning for Engineering (TALE)},
  pages     = {1--8},
  year      = {2025},
  doi       = {10.1109/tale66047.2025.11346682}
}

@article{ke2025retrieval,
  author  = {Ke, Yuhe and Jin, Liyuan and Elangovan, Kabilan and Abdullah, Hairil and Liu, Nan and Sia, Alex Tiong Heng and Soh, Chai Rick and Tung, Joshua Yi Min and Ong, Jasmine and Kuo, Chun-Ju and Wu, Shao-Chun and Kovacheva, Vesela and Ting, Daniel},
  title   = {Retrieval Augmented Generation for 10 Large Language Models and its Generalizability in Assessing Medical Fitness},
  journal = {npj Digital Medicine},
  volume  = {8},
  year    = {2025},
  doi     = {10.1038/s41746-025-01519-z}
}

@article{levonian2023retrieval,
  author  = {Levonian, Zachary and Li, Chenglu and Zhu, Wangda and Gade, Anoushka and Henkel, Owen and Postle, Millie-Ellen and Xing, Wanli},
  title   = {Retrieval-Augmented Generation to Improve Math Question-Answering: Trade-offs Between Groundedness and Human Preference},
  journal = {arXiv preprint arXiv:2310.03184},
  year    = {2023},
  doi     = {10.48550/arxiv.2310.03184}
}

@article{li2024enhancing,
  author  = {Li, Jiarui and Yuan, Ye and Zhang, Zehua},
  title   = {Enhancing {LLM} Factual Accuracy with {RAG} to Counter Hallucinations: A Case Study on Domain-Specific Queries in Private Knowledge-Bases},
  journal = {arXiv preprint arXiv:2403.10446},
  year    = {2024},
  doi     = {10.48550/arxiv.2403.10446}
}

@article{li2025retrieval,
  author  = {Li, Zhixun and Wang, Zhen and Wang, Wei and Hung, Kevin and Xie, Haoran and Wang, Fu Lee},
  title   = {Retrieval-Augmented Generation for Educational Application: A Systematic Survey},
  journal = {Computers and Education: Artificial Intelligence},
  volume  = {8},
  pages   = {100417},
  year    = {2025},
  doi     = {10.1016/j.caeai.2025.100417}
}

@article{linders2025knowledge,
  author  = {Linders, Jasper and Tomczak, Jakub M.},
  title   = {Knowledge Graph-Extended Retrieval Augmented Generation for Question Answering},
  journal = {Applied Intelligence},
  volume  = {55},
  year    = {2025},
  doi     = {10.1007/s10489-025-06885-5}
}

@article{lyu2024crudrag,
  author  = {Lyu, Yuanjie and Li, Zhiyu and Niu, Simin and Xiong, Feiyu and Tang, Bo and Wang, Wenjin and Wu, Hao and Liu, Huanyong and Xu, Tong and Chen, Enhong},
  title   = {{CRUD-RAG}: A Comprehensive {Chinese} Benchmark for Retrieval-Augmented Generation of Large Language Models},
  journal = {ACM Transactions on Information Systems},
  volume  = {43},
  pages   = {1--32},
  year    = {2024},
  doi     = {10.1145/3701228}
}

@inproceedings{maity2024leveraging,
  author    = {Maity, Subhankar and Deroy, Aniket and Sarkar, Sudeshna},
  title     = {Leveraging In-Context Learning and Retrieval-Augmented Generation for Automatic Question Generation in Educational Domains},
  booktitle = {Proceedings of the 16th Annual Meeting of the Forum for Information Retrieval Evaluation},
  year      = {2024},
  doi       = {10.1145/3734947.3734949}
}

@article{neha2025retrieval,
  author  = {Neha, Fnu and Bhati, Deepshikha and Shukla, Deepak Kumar},
  title   = {Retrieval-Augmented Generation ({RAG}) in Healthcare: A Comprehensive Review},
  journal = {AI},
  volume  = {6},
  number  = {9},
  year    = {2025},
  doi     = {10.3390/ai6090226}
}

@article{peng2024graph,
  author  = {Peng, Boci and Zhu, Yun and Liu, Yongchao and Bo, Xiaohe and Shi, Haizhou and Hong, Chuntao and Zhang, Yan and Tang, Siliang},
  title   = {Graph Retrieval-Augmented Generation: A Survey},
  journal = {ACM Transactions on Information Systems},
  volume  = {44},
  pages   = {1--52},
  year    = {2024},
  doi     = {10.1145/3777378}
}

@article{wright2026multimodal,
  author  = {Wright, Brian and others},
  title   = {Using Educational Data to Explore Multimodal (Audio, Visual, and Textual) {LLM} Retrieval Techniques},
  journal = {Working paper},
  year    = {2026},
  note    = {Prior study; full venue details to be added}
}

@article{xu2025megarag,
  author  = {Xu, Shungeng and Yan, Zhenghan and Dai, Chengzhi and Wu, Fan},
  title   = {{MEGA-RAG}: A Retrieval-Augmented Generation Framework with Multi-Evidence Guided Answer Refinement for Mitigating Hallucinations of {LLMs} in Public Health},
  journal = {Frontiers in Public Health},
  volume  = {13},
  year    = {2025},
  doi     = {10.3389/fpubh.2025.1635381}
}

@article{zhou2025indepth,
  author  = {Zhou, Yingli and Su, Yaodong and Sun, Youran and Wang, Shu and Wang, Taotao and He, Runyuan and Zhang, Yongwei and Liang, Sicong and Liu, Xilin and Fang, Yuchi},
  title   = {In-depth Analysis of Graph-based {RAG} in a Unified Framework},
  journal = {arXiv preprint arXiv:2503.04338},
  year    = {2025},
  doi     = {10.48550/arxiv.2503.04338}
}

@article{zhu2025graphsurvey,
  author  = {Zhu, Zulun and Huang, Tiancheng and Wang, Kai and Ye, Junda and Chen, Xinghe and Luo, Siqiang},
  title   = {Graph-Based Approaches and Functionalities in Retrieval-Augmented Generation: A Comprehensive Survey},
  journal = {ACM Computing Surveys},
  year    = {2025},
  doi     = {10.1145/3795880}
}

@inproceedings{zhu2025kg2rag,
  author    = {Zhu, Xiangrong and Xie, Yuexiang and Liu, Yi and Li, Yaliang and Hu, Wei},
  title     = {Knowledge Graph-Guided Retrieval Augmented Generation},
  booktitle = {Proceedings of the 2025 Conference of the North American Chapter of the Association for Computational Linguistics},
  pages     = {8912--8924},
  year      = {2025},
  doi       = {10.48550/arxiv.2502.06864}
}

@article{kestin2025tutoring,
  author  = {Kestin, Greg and Miller, Kelly and Klales, Anna and Milbourne, Timothy and Ponti, Gregorio},
  title   = {{AI} Tutoring Outperforms In-Class Active Learning: An {RCT} Introducing a Novel Research-Based Design in an Authentic Educational Setting},
  journal = {Scientific Reports},
  volume  = {15},
  year    = {2025},
  doi     = {10.1038/s41598-025-97652-6}
}

@inproceedings{ko2026toward,
  author    = {Ko, Eunhye Grace and Lee, Hakeoung Hannah and Singh, Anjali and Boddy, Lily and Ford, Kasey and Huff, Earl W.},
  title     = {Toward Scalable and Responsible Integration of Course-Specific {AI} Tutors: Instructor Experiences with a Campus-Wide Platform},
  booktitle = {Proceedings of the 2026 CHI Conference on Human Factors in Computing Systems},
  year      = {2026},
  doi       = {10.1145/3772318.3791298}
}

@article{li2025teacher,
  author  = {Li, Yan and Wu, Yuanyuan and Chiu, Thomas K. F.},
  title   = {How Teacher Presence Affects Student Engagement with a Generative Artificial Intelligence Chatbot in Learning Designed with First Principles of Instruction},
  journal = {Journal of Research on Technology in Education},
  volume  = {58},
  number  = {5},
  year    = {2025},
  doi     = {10.1080/15391523.2025.2493942}
}

@article{neagu2026rethinking,
  author  = {Neagu, Alexandra and Wong, Jeffrey T. H. and Messer, Marcus and Nelson, Rhodri and Johnson, Peter B.},
  title   = {Rethinking Scaffolding in {LLM} Tutors: The Interactional Mismatch Between Benchmarks and Real-World Deployments},
  journal = {arXiv preprint arXiv:2606.15766},
  year    = {2026},
  doi     = {10.48550/arxiv.2606.15766}
}

@article{sunil2025socraticai,
  author  = {Sunil, Karthik and Thakkar, Aalok},
  title   = {{SocraticAI}: Transforming {LLMs} into Guided {CS} Tutors Through Scaffolded Interaction},
  journal = {arXiv preprint arXiv:2512.03501},
  year    = {2025},
  doi     = {10.48550/arxiv.2512.03501}
}

@article{thway2025harnessing,
  author  = {Thway, Maung and Recatala-Gomez, Jose and Lim, Fun Siong and Hippalgaonkar, Kedar and Ng, Leonard W. T.},
  title   = {Harnessing {GenAI} for Higher Education: A Study of a Retrieval Augmented Generation Chatbot's Impact on Learning},
  journal = {Journal of Chemical Education},
  volume  = {102},
  number  = {9},
  year    = {2025},
  doi     = {10.1021/acs.jchemed.5c00113}
}

@article{wall2025generative,
  author  = {Wall, Vince and Bedford, Alison and Redmond, Petrea},
  title   = {Generative Artificial Intelligence in Education: Initial Principles Developed from Practitioner Reflexive Research},
  journal = {The Journal of Educational Research},
  volume  = {118},
  number  = {6},
  year    = {2025},
  doi     = {10.1080/00220671.2025.2510398}
}

@article{wang2025generative,
  author  = {Wang, Xiaoyu and Zainuddin, Zamzami and Leng, Chin Hai},
  title   = {Generative Artificial Intelligence in Pedagogical Practices: A Systematic Review of Empirical Studies (2022--2024)},
  journal = {Cogent Education},
  volume  = {12},
  number  = {1},
  year    = {2025},
  doi     = {10.1080/2331186x.2025.2485499}
}

@article{wu2024chatbots,
  author  = {Wu, Rong and Yu, Zhonggen},
  title   = {Do {AI} Chatbots Improve Students' Learning Outcomes? Evidence from a Meta-Analysis},
  journal = {British Journal of Educational Technology},
  volume  = {55},
  number  = {1},
  year    = {2024},
  doi     = {10.1111/bjet.13334}
}

@article{wu2026chatgpt,
  author  = {Wu, Xinning and Zhu, Pei and Zhang, Jinliang and Yin, Mengwei and Wang, Yingxi},
  title   = {{ChatGPT}'s Impact on Student Learning Outcomes: A Meta-Analysis of 35 Experimental Studies},
  journal = {Humanities and Social Sciences Communications},
  volume  = {13},
  number  = {1},
  year    = {2026},
  doi     = {10.1057/s41599-026-07019-z}
}

@article{cepeda2006distributed,
  author  = {Cepeda, Nicholas J. and Pashler, Harold and Vul, Edward and Wixted, John T. and Rohrer, Doug},
  title   = {Distributed Practice in Verbal Recall Tasks: A Review and Quantitative Synthesis},
  journal = {Psychological Bulletin},
  volume  = {132},
  number  = {3},
  pages   = {354--380},
  year    = {2006},
  doi     = {10.1037/0033-2909.132.3.354}
}

@article{karpicke2008critical,
  author  = {Karpicke, Jeffrey D. and Roediger, Henry L.},
  title   = {The Critical Importance of Retrieval for Learning},
  journal = {Science},
  volume  = {319},
  number  = {5865},
  pages   = {966--968},
  year    = {2008},
  doi     = {10.1126/science.1152408}
}

@article{rohrer2007shuffling,
  author  = {Rohrer, Doug and Taylor, Kelli},
  title   = {The Shuffling of Mathematics Problems Improves Learning},
  journal = {Instructional Science},
  volume  = {35},
  number  = {6},
  pages   = {481--498},
  year    = {2007},
  doi     = {10.1007/s11251-007-9015-8}
}

@inproceedings{lewis2020retrieval,
  title={Retrieval-augmented generation for knowledge-intensive NLP tasks},
  author={Lewis, Patrick and Perez, Ethan and Piktus, Aleksandra and Petroni, Fabio and Karpukhin, Vladimir and Goyal, Naman and Kulikov, Igor and Ghazvininejad, Marjan and Zettlemoyer, Luke and Kiela, Douwe},
  booktitle={Advances in Neural Information Processing Systems},
  volume={33},
  pages={9459--9474},
  year={2020}
}

\end{document}